\documentclass[letterpaper, 10 pt, conference]{ieeeconf}  

\IEEEoverridecommandlockouts                              

\usepackage[
  backend=biber,
  style=ieee,
  citestyle=numeric-comp, 
  autocite=superscript,    
  sortcites=true          
]{biblatex}

\usepackage[version=4]{mhchem}
\usepackage{siunitx}
\usepackage{grffile}
\usepackage{stfloats}

\title{\LARGE \bf
Functionalization of Situated Robots via Vapour*
}

\author{Kadri-Ann Pankratov$^{1}$, Leonid Zinatullin$^{1}$, Adele Metsniit$^{1}$, Marie Vihmar$^{1}$, and Indrek Must$^{1}$
\thanks{*This work was supported by Estonian Research Council Grant PRG1498, Kristjan Jaak Scholarship for Short Study Visits, Estonian Doctoral School mobility Grant (funded by the European Union)}
\thanks{$^{1}$ authors are with Institute of Technology, University of Tartu, Nooruse 1, 50411 Tartu, Estonia, 
        {\tt\small kadri-ann.pankratov@ut.ee, indrek.must@ut.ee}}%
}

\begin{document}

\maketitle
\thispagestyle{empty}
\pagestyle{empty}

\begin{abstract}

In this Extended Abstract, we show recent unpublished developments on functionalizing in situ built robotic bodies with environmental materials.

\end{abstract}

\section{INTRODUCTION}

Robot operation in complex environments benefits from tight matching with the environment for functionality. We know how to spin robot embodiments \cite{Vihmar2025Silk-inspiredRobots, Vihmar2026FibrousEnvironments}, such as passive/subordinate grippers in situ according to exact demands, but functionalization remains an integration challenge. This could be solved by combining active materials via multimaterial spinning which requires complex spinneret multiplexing, reducing operability in environment. Alternatively, spinning mixture doping remains limited to available additives and chemical stability during spinning. Using available materials from the environment reduces payload and the in-situ developed structures are uniqely matched with environmental allowances, a strong advantage in front of prefabrication.

In this work, we suggest a pathway in which the robot first deploys a simple, non-fucntionalized fiber to build a structure, and then the web is functionalized as a result of fiber-environment interaction (Fig. A). We demonstrate transformation of an easily spun optically scattering polymer fiber into an optically absorbing, polypyrrole-grafted web, as an example of modulating optical properties. The functionalization activator is delivered to the web by liquid or embedded in the spinning mixture, highlighting two possible strategies of preparedness for environmental conditions (here pyrrole). Here we demonstrate the concept in an engineered environment for feasibility, but we foresee other relevant environmentally specific elements, such as bacterial genomes for specific biomolecule synthesis in a biohybrid robot, as potential applications.

\section{Methods}

PVDF-webs were prepared according to our previous work in \cite{Vihmar2025Silk-inspiredRobots} (Fig. 1B). As an activator (oxidizing agent), we used \ce{FeCl3} (Czysty pure), similarly to \cite{Mohammadi1986ChemicalPolypyrrole}. The activator was introduced in two separate strategies: via infusion of a activator-carrying liquid (approximately 5\% w\% mixture in propylene carbonate (Alfa Aesar) or by embedding the activator in the web mixture (by adding approximately \SI{1}{gram} of \ce{FeCl3}). To enable liquid transport from the support to the web, we used porous Teflon tube (ePTFT, diameter \SI{1}{\milli\meter}, Fluoroplastics Store) for the infused activator. A 3D printed (polyethylene terephtalate glycol (PETG), Bambu Lab A1) scaffold, inspired by the vein patterns in \textit{Catocala fraxini} was used for support of the web with the embedded activator.

Pyrrole (Py, Sigma, distilled under reduced pressure and stored under argon at \SI{-18}{\degree}C) environment was simulated by dropping a few drops of Py onto a glass Petri dish on a hotplate at approximately \SI{60}{\degree}C to evaporate the Py (chemical vapour deposition, Fig. 1C). The substrate was held in the vapours for 1-2 minutes, until the the substrate had turned into PPy (evident via colour change to black). 

SEM imaging was performed with Hitachi TM3000 scanning electron microscope with back-scatter electron detector and 15-kV acceleration voltage. Canon EOS 60D was used for photography. Background in Fig. 1D (left) and E (left) was removed with Photoshop.

LLMs were used for clarity and language.

\section{Results}

Functionalisation of the PVDF webs was achieved using two complementary activator‑delivery strategies: (i) liquid‑infusion of activator into a prefabricated web (Fig. 1D), and (ii) activator pre‑embedding within the spinning mixture prior to web formation (Fig. 1E).

In the liquid‑infusion approach, the \ce{FeCl3} solution was introduced through the porous Teflon support, allowing it to wick into and wet the web. Upon exposure to pyrrole vapour, PPy polymerised preferentially on the liquid film present on the fibers. Regions with greater fluid accumulation produced a continuous, fiber‑reinforced PPy membrane reminiscent of butterfly wing membrane (Fig 1D, right top), whereas drier regions developed a conformal PPy coating around the fibres (Fig 1D, right bottom). This spatial gradient in morphology demonstrates that liquid‑mediated activation can be locally tuned, enabling the web to selectively capture and convert environmental compounds into functional structural elements.

In the embedded activator approach, the activator was mixed directly into the spinning solution before blow‑spinning. These pre‑loaded webs underwent successful PPy polymerisation under the same vapour‑phase conditions. SEM and optical inspection revealed PPy‑PVDF formations concentrated in web nodes and droplets (Fig. 1, right) where activator content was higher, confirming that the embedded activator remained active and accessible. This supports the concept of incorporating reactive "chemical placeholders" into the web architecture, enabling later activation by environmental vapours without requiring additional liquid infusion.

Together, these results establish that both wetting‑based activation and embedded activator loading can autonomously drive in situ formation of thin‑film or fiber‑level PPy structures. The functionalization process in both cases depends on the amount of environmental material (Py vapour). These behaviours highlight how artificial webs can be engineered to gather, bind, and transform environmental materials into functional extensions of a robot’s body.

\begin{figure*}[!t]
    \centering
    \includegraphics[width=\textwidth]{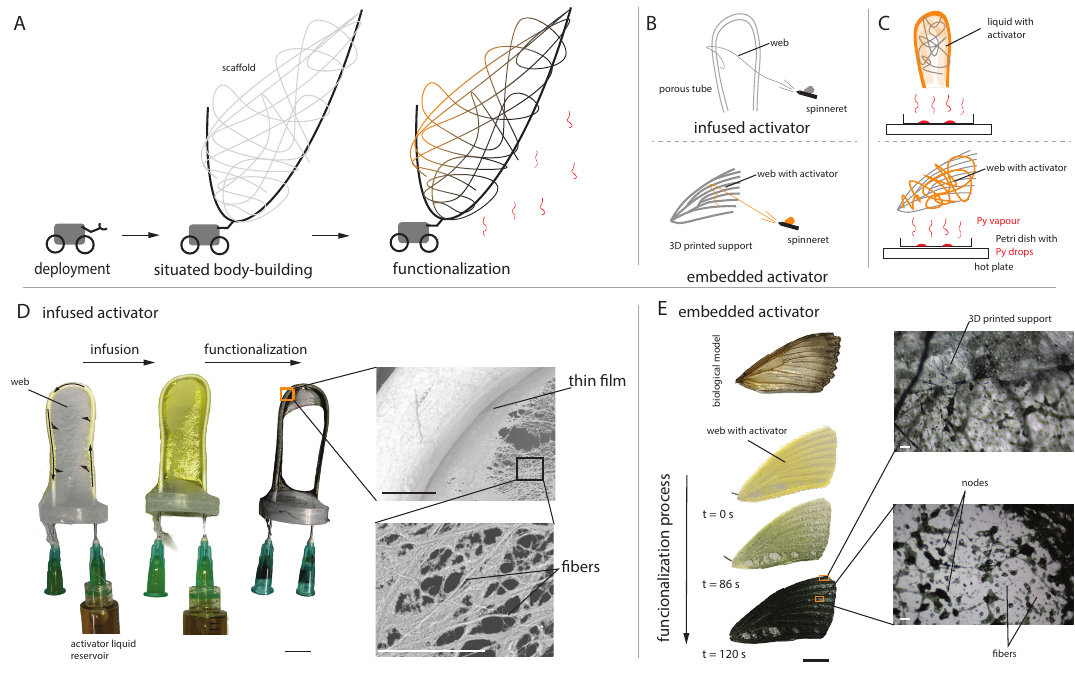}
    \caption{Fig 1. A, Situtated body functionalization concept. B, Spinning on supports: porous tube for activator introduction via infusion (top), and 3D printed bioinspired support for embedded activator in the web mixture. C, Scaffold activation in a Py vapour environment. D, Infusion and functionalisation of web on the porous tube. E, Functionalization process of the web with embedded activator. Scale bars: (D- left, E-left) - \SI{1}{\centi\meter}, (D - right top) - \SI{1}{\milli\meter}, (D-right bottom, E - right) - \SI{100}{\micro\meter}}
    \label{fig}
\end{figure*}

\section{CONCLUSIONS}

In conclusion, we showed functionalization of situated robotic scaffolds by polymerising material from the environment. For soft robotics, this demonstrates a practical pathway for robots to harvest, synthesise, and integrate new body components in situ, supporting adaptive morphologies, self‑reinforcement, and environmentally driven body evolution. Future work could focus on tailoring the properties of PPy for situated large-area applications, such as solar powered water evaporation \cite{Wang2018ImprovedCones}, bioscaffolding \cite{Liang2021ConductiveEngineering}, and humidity sensors \cite{Hussain2024Polypyrrole:Sensor}.

\addtolength{\textheight}{-4cm}   




\section*{ACKNOWLEDGMENT}

The Authors thank Dr. Edoardo Sinibaldi (Italian Institute of Technology) for scientific discussions and colleagues Dr. Toomas Tammaru and Kadri Ude (University of Tartu) for freeze-dried \textit{Catocala fraxini} samples.


\printbibliography

\end{document}